# Coupling Learning of Complex Interactions

Longbing Cao

*Advanced Analytics Institute, University of Technology Sydney, Australia*
*Telephone: +61-2-95144477, Email: longbing.cao@uts.edu.au.*

## Abstract

Complex applications such as big data analytics involve different forms of coupling relationships that reflect interactions between factors related to technical, business (domain-specific) and environmental (including socio-cultural and economic) aspects. There are diverse forms of couplings embedded in poor-structured and ill-structured data. Such couplings are ubiquitous, implicit and/or explicit, objective and/or subjective, heterogeneous and/or homogeneous, presenting complexities to existing learning systems in statistics, mathematics and computer sciences, such as typical dependency, association and correlation relationships. Modeling and learning such couplings thus is fundamental but challenging. This paper discusses the concept of *coupling learning*, focusing on the involvement of coupling relationships in learning systems. Coupling learning has great potential for building a deep understanding of the essence of business problems and handling challenges that have not been addressed well by existing learning theories and tools. This argument is verified by several case studies on coupling learning, including handling coupling in recommender systems, incorporating couplings into coupled clustering, coupling document clustering, coupled recommender algorithms and coupled behavior analysis for groups.

*Keywords:* Coupling learning, coupling relationship, interaction learning, non-IIDness learning, invisibility learning

## 1. Introduction

Complex interactive and unstructured/semi-structured data and applications, especially in big data, present major challenges to the current analytic and learning theories and systems. Big data, in particular, presents specific complexities of weakly structured and unstructured data distribution, dynamics, interactions, and structures, which challenge the existing theoretical and commercial systems in mathematics, statistics, and computer science. Examples include the connections between gene combinations and physical and psychological consequences, between one's personal traits or preferences in social media and one's social, behavioral, attitudinal and interest attributes.

This results in a situation where learning big data is analogous to the ancient Indian parable of seven blind men encountering an elephant for the first time. Each touches a different part of the animal, so when the seven share their experiences, each has a completely different idea of what the whole animal must look like. Similarly, when confronted with a big data set, a data modeler or learner may only see a partial set or aspect, hence often only a partial story is told by a learner. Why does this happen? There are many reasons, one of which is the invisibility of sophisticated coupling relationships (coupling for short, see Definition 2.1) hidden between the heterogeneous parts that are 'visible' to blind people. They do not have the ability to recognize the visible and invisible couplings between parts to connect those heterogeneous parts to form a global picture as sighted people do. This is representative of certain major challenges of complex relations hidden in complex data (particularly referring here to data with complex couplings and/or mixed distributions, formats, types and variables, and unstructured and weakly structured data). Learning visible and especially invisible coupling relationships can complement and assist in understanding weakly structured and unstructured data.

In many cases, such inherent, locally visible but globally invisible (or vice versa) couplings are presented in a range of forms, structures, and layers and on diverse entities. Often individual learners cannot tell the whole story due to their inability to identify such complex coupling. Effectively learning the widespread, various, visible and invisible couplings is thus crucial for obtaining a true and total picture of the underlying problem.

This is not a trivial task, however. The difficulty in learning complex couplings lies not only with invisible couplings - even *visible couplings* are often overlooked. Taking the design of recommender algorithms as an example, our ability to recognize them is limited, even though these interactions and structures are embedded in applications such as social media networks. For example, there have been several recent cases in which researchers have started to incorporate inherent couplings between items and between users into a recommender system (RS) [49, 28], after a long period of focusing on rating-based exploration, whereas the item-item couplings and user-user couplings (see Figure 2) have been always intrinsic to the systems.

One reason for this is that visibility is relative to opportunity and capability. The same couplings are implicit to some people, while explicit to others. For instance, in social media recommendation, the friendship between twitters [16] has only recently been recognized as enhancing social recommendation, yet it has always been a natural built- in feature of social media systems. There is a need to develop our ability to capture and convert as many invisible couplings as possible to visible coupling, and to effectively capture visible couplings in complex data.

In reviewing the existing literature, we unfortunately cannot find systematic methodologies and techniques in learning theories to address the above coupling issues. This raises a fundamental question: *how much do we know about coupling?* and many other basic questions, including: *what are couplings*, *where they are*, and *in what forms are they present*, which we need to address before we can think about how to capture and embed couplings in learning systems. Once these problems have been satisfactorily addressed, more issues follow, such as: *how to represent couplings*, *how to test whether and to what extent couplings exist in a dataset*, *how to incorporate them into learning models*, and *how to evaluate the difference they make once they are incorporated into learning systems*. These challenges form the basis of the need to study *coupling learning*, a fundamental but undeveloped area in computer science, to address the intricate coupling relationships embedded in complex data and increasingly seen in information retrieval, data mining and machine learning in particular. This is crucial for big data analytics because most existing analytics and learning theories and systems have been built on the assumption that data is independent and identically distributed (IID), while big data is essentially non-IID [6]. Coupling is one critical aspect of non-IIDness [6] (the other is heterogeneity or so-called personalization, which is not the main concern in this paper, although coupling may be heterogeneous and involve heterogeneity in data).

Learning the above characteristics of complex couplings in big data fundamentally challenges existing learning theories and systems, including pairwise coupling [37, 67], statistical relation learning [21, 22], dependency learning [44, 64], association learning [15, 34], correlation analysis [26, 53], linkage analysis [25, 36], community analysis [2, 23], social network analysis [2, 23, 30, 63], multivariate time series [53], causality analysis [24] and graph analysis [18]. They either essentially treat data as IID or only address specific forms or levels of couplings. No general and competent theories, frameworks, algorithms or tools are available to handle the coupling complexities discussed above.

The above observations motivate this work, namely to systematically state the coupling learning problem, which clearly involves interactive, unstructured and semi-structured data. The aim of this paper is multi-fold:

- High-level: build a conceptual system of coupling learning (Sections 2-4) towards a generic and comprehensive understanding of the broad-based coupling relationships that exist in complex data and applications (especially in big data related business);

- Middle-level: illustrate how to advance classic problems to another generation by incorporating coupling learning into a specific existing scientific problem such as recommender systems (Section 5), and

- Low-level: showcase specific examples in recommender systems to demonstrate how couplings can be managed in practice to improve analytic outcomes (Section 6).

The purpose of this paper is therefore not to specify one particular technique for learning a particular type of coupling (instead we provide citations to our related work for such discrete discussions), but to disclose the whole nature of the problem and build generic frameworks and examples to show possible ways to address the problem.

Accordingly, the organization of this work is as follows. Section 2 discusses the concept of coupling and major coupling relationships often addressed in current big data communities. Section 3 presents a high-level picture of coupling layers and forms appearing in complex data and applications. In Section 4, the issues of modeling and measuring couplings and the curse of couplings are introduced. An example of comprehensive couplings in recommender systems is discussed in Section 5, which presents a new theoretical framework for next-generation recommender systems. Two case studies are given in Section 6, one in which a coupled K-mode algorithm to identify items with strong coupling relationships is presented, and one in which couplings are utilized to improve Matrix Factorization-based recommendation. Section 7 explores the opportunities for learning couplings in data mining, text mining, information retrieval, and complex behavior analysis. The paper is concluded in Section 8.

## 2. Coupling: An important perspective

In this section, we discuss the concept of coupling, and the relevant work in statistics, mathematics and computer science. The following key concepts are used in this paper:

- Coupling: refers to any relationship or interaction that connects two or more aspects (which could be between inputs or between inputs and outputs);

- Aspect: a term broadly referring to entity, entity property (or characteristics such as variations), property value, context, learner or analytic model, learning objective (effect), or learned results (such as patterns and clusters), etc. Typically, an aspect refers to either an antecedent or a postcedent, or to either a cause or an effect.

- Entity: refers to elements such as profile (configuration), object class (category), object, instance, etc. on which learning is applied; in a practical problem such as recommender systems, entity is instantiated to respective elements, such as user, item, and rating;

- Context: refers to the surrounding environment and environmental elements;

- Property: refers to the characteristics of an entity and is interchangeable with attribute and variable;

- Interaction: broadly refers to invisible or visible connections that exist between two or more entities;

- Intra-coupling: refers to the coupling within an aspect, such as the relationship between the values of a property;

- Inter-coupling: refers to coupling between aspects, such as between properties;

- Coupling layer: refers to where the coupling lodges, such as entity, property, context, interaction and learning;

- Coupling form: refers to the way in which coupling is presented, in terms of specific presentations, configurations, profiles, structures, types, and/or variations of coupling.

- Coupling strength: refers to the extent to which the connected aspects are bonded. Strengths may be represented in terms of frequency, similarity, dissimilarity, mutual information, probability, likelihood, effect (such as effectiveness, efficiency), business values (such as economic performance between financial variables) or other metrics. The strength of couplings between two bonded aspects is sensitive to the coherence (or compactness) of its constituent forms (configurations).

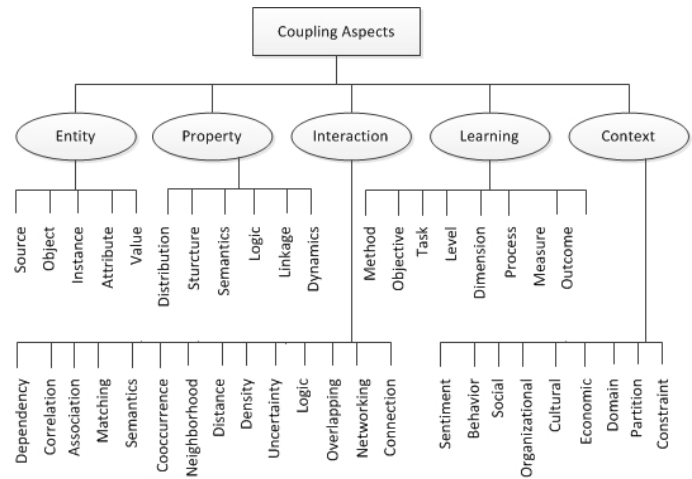

Figure 1. Coupling aspects.

## 2.1. The concept of coupling

Coupling is a very rich concept, related to many aspect as discussed before and below.

**Definition 2.1.** *(Coupling) Coupling refers to any relationship (for instance, co-occurrence, neighborhood, dependency, linkage, correlation, or causality) between two or more aspects, such as object, object class, object property (variable), process, fact and state of affairs, or other types of entities or properties (such as learners and learned results) appearing or produced prior to, during and after a target process (such as a learning task).*

For example, different couplings exist between ratings for an item, items, item properties, users and user properties in a recommender system, as shown in Figure 2 in Section 5.

In a learning system, as shown in Figure 1, couplings may exist within and/or between aspects, such as entity (objects, object class, instance, or group/community) and its/their properties (variables), context (environment) and its constraints, interactions (exchange of information, material or energy) between entities or between the entity and its/their environment, learning objectives (targets, such as risk level or fraud), the corresponding learning methods (models, algorithms or systems) and resultant outcomes (such as patterns or clusters).

The process of learning couplings further requires the following perspectives to be considered: problem domain, data presentation (format), knowledge representation, and coupling characteristics.

- From the problem domain perspective, coupling may present in terms of domain-specific, cross-domain, organizational, social, behavioural, cultural, or economic connections, depending on which specific areas a problem belongs to. For example, in the economic domain, different markets affect each other. Price movement in one stock market may be influenced by prices in other stock markets and other types of markets such as petrol prices, commodity prices, and exchange rates.

- On the data format side, coupling may appear in different forms corresponding to diverse data formats, including numerical, categorical, textual, graphic, image, or mixed-structured data. For instance, in [8] and [59] respectively, behavior coupling and categorical coupling between objects are discussed. From an even higher level perspective, coupling may exist between entity profiles (configurations), which may take the form of arrays, affinities, artworks, or networks.

- In terms of knowledge representation, coupling may be presented in terms of syntactic, semantic (ontological), graphic, or inferential specifications. For example, in [58, 5], different temporal logic-based, ontological and inferential couplings are discussed for coupled group behaviors. In a mathematical way, couplings may be presented as configurations, distributions, linkage, neighboring, co-occurrences, causality, correlations and dependence.

. Couplings may appear to have different characteristics: explicit (visible) vs. implicit (invisible), qualitative (e.g. social coupling) vs. quantitative (e.g. conditional dependency), descriptive analysis vs. deep analysis, specific (e.g. rating similarity in RS) vs. comprehensive (e.g. item and user-coupling similarity in RS), local (part) vs. global (body), certain (static) vs. uncertain (dynamic), and known (e.g. distance calculation) vs. unknown (latent relation), according to the characteristics of the aspects involved in a system. In addition to typical relationships such as correlations between two variables and causality between inputs and outputs, couplings may present characteristics of Simpson's Paradox [62] or Granger Causality [24].

## 2.2. Classic coupling relationships

The above discussions show that coupling as a concept covers, but is also much richer than, the existing terms including dependence [44, 64], correlation [26, 53], and association [15, 34], and involves similarity learning [45]. Classic research on dependence, correlation and association are much more specific, descriptive and explicit, often having specific meaning and ways of being measured. Similarity learning serves a fundamental purpose in object relation analysis. In this section, we review these typical coupling relationships which are mainly addressed in statistics, mathematics, and computer science.

In statistics and mathematics, typical coupling relationships include association, correlation and dependence. *Association* refers to any relationship between two measured quantities that renders them statistically dependent [57]. *Correlation* [26, 53] is a concept with a narrower scope and more specific meaning, usually referring to a linear relationship between two quantities. *Dependence* [44, 64] refers to an even more specific statistical relationship between two random variables or two sets of data which do not satisfy a mathematical condition of probabilistic independence.

Many statistical and mathematical measures have been proposed to quantify the extent of association, correlation and dependence. Examples of such measures include the product moment correlation coefficient, the odds ratio used for dichotomous measurements, the distance correlation, tetrachoric correlation coefficient, Goodman and Kruskal's lambda, and Tschuprow's T and Cramer's V. There are several correlation coefficients available for measuring the degree of correlation in very specific cases, such as Pearson correlation coefficient. Other correlation measures have been introduced to address their deficiency, such as distance correlation and Brownian correlation [55, 56], the correlation ratio to detect functional dependence, total correlation and dual total correlation to detect general dependence, multi-moment correlation measures, polychoric correlation for latent ordinal variables, and copula for connecting dependence between variables.

Specific focus has been placed on correlation and linearity [53, 50], and correlation and causality [40] in statistics and machine learning. Linear correlation, such as Pearson correlation coefficient and canonical correlation analysis, has been widely explored. Nonlinear relationships [19] and correlation with the variation of the correlation coefficient with time have been less explored but appear widely in real-world data and applications. The conventional dictum "correlation does not imply causation" indicates that the causes underlying correlation between variables may be direct (known) or indirect (unknown). Causality (causation) [46] (such as Granger Causality [24]) indicates the reasons driving one event after another. Typical theory includes probabilistic causation which measures the likelihood of the cause triggering effect. However, couplings may go far beyond causality, such as implicit connections, indication or co-occurrences.

The specific meanings of the mathematical/statistical terms *association*, *correlation* and *dependence* have been widened to satisfy needs in other domains. Typically, association is a common concept widely used in various fields of study, for example, astronomical associations between astronomical exposures, archaeological associations between objects found together, chemical associations or dissociations between particles, ions or radicals, ecological associations composing species, genetic associations (linkage, or occurrences) of two or more traits in a population of individuals, and psychological associations showing connections between two or more concepts in the mind or imagination.

In computer science, there is a long history of research on specific coupling relationships. In software engineering, coupling (grouping) is a concept widely accepted in computer programming which measures the degree to which each program module relies on each one of the other modules. There, coupling is usually contrasted with cohesion. Low coupling is often a sign of a well-structured computer system and a good design [73]. Different levels of couplings exist in a software system, from data, message, module to communication and control. Such software couplings do not involve analytic and learning tasks, thus they are not discussed in this paper.

Researchers in the areas of information theory, data mining, machine learning, artificial intelligence, pattern recognition and image processing have specifically paid significant attention to similarity learning, which serves as a fundamental tool for measuring and analyzing object relationships. Many similarity measures have been proposed for different data presentations, which address object coupling in terms of major paradigms, including co-occurrence, coverage (overlapping), distance (difference or dissimilarity), causality and uncertainty. Most of the effort has focused on addressing strong (also explicit, high or tight) couplings or decouplings (independence) as a priority. Low (also implicit, loose or weak) couplings are often ignored to simplify the analytic complexity.

Additional efforts have been made in information theory [48], fuzzy set theory [29, 35] and statistical machine learning [27] to address uncertainty or possibility and build respective entropy theories (through entropy-based mutual information, for instance), fuzzy membership functions and probabilistic similarity. In recent years, latent relation analysis [20, 8, 6] has been increasingly picked up by the machine learning community, while the existing work still focuses on specific couplings, especially probabilistic relationships. There are very few studies on other types of couplings, which are more related to non-entity couplings (such as multi-source, method and feature couplings) and other types of entities or properties appearing or produced prior to, during and after a target process (such as a learning task). No systematic discussions can be found on coupling learning from the problem domain, data presentation, knowledge representation, and coupling characteristic perspectives.

## 3. Ubiquitous couplings

In this section, we expand the above discussions on couplings, aiming to provide an overall picture of couplings widespread in comprehensive learning tasks.

### 3.1. Coupling layers

As discussed in Section 2.1, the complexity of couplings is embodied through many aspects related to *entity*, *property*, *context*, *interaction* and *learning* in a learning system. Accordingly, we here discuss the couplings present in these five layers.

- Entity coupling: An entity is the basic carrier of couplings. Couplings may exist within and/or between data sources, data objects, instances, properties, and values. Correspondingly, intra-coupling occurs within an entity, such as intra-attribute coupling [59], and inter-coupling appears between entities, for example, inter-attribute coupling [59]. Couplings associated with different data presentations (different formats and types) of entities present in different forms. Couplings in numeric data [61] are very different from those in categorical data [59]. Of the total number of entities involved, such as attributes, couplings include single entity-based couplings such as temporal coupling, and compound couplings (which involve multiple couplings organized in a certain way or structure) such as hierarchical coupling (for more about behavior combinations, refer to [8, 58] and for pattern relations, refer to [5, 11]).

- Property coupling: Property (attribute) coupling may take place in terms of entity property distribution, structure, semantics, logic, linkage or dynamics. Intra-property coupling refers to the coupling between the values of a property, such as age coupling. In addition, one property of an entity may more or less depend on, or influence, another property, which involves inter-property coupling between properties. For example, intra-term and inter-term coupling metrics are defined in [17] to disclose the semantic linkage between terms in document analysis, and intra- and inter-attribute coupling are discussed in [59]. In addition to the traditional focus on the coefficient between numerical variables, more work is needed to explore the interactions between heterogeneous properties, such as those found in multi-structured data.

- Context coupling: The context surrounding an underlying learning problem may be composed of aspects supporting or influencing entity properties, interactions and effect. This may include a mixture of varied factors in the broader context, from social, organizational, cultural, economic, and domain-specific aspects, besides the specific constraints surrounding entities and systems. In addition to this, entities interact with their context, and corresponding couplings also exist within and between these aspects. Capturing the couplings between entities and their contexts may occur through latent learning (see the discussions about user-item couplings in Figure 2 in Section 5 [6]).

- Interaction coupling: From the interaction channel perspective, couplings may take the form of dependence, correlation, association, matching, semantic similarity, co-occurrence, neighbourhood, distance, density, uncertainty, logic relation, overlapping, networking or connection. In a complex application, such interactions may occur within and between system elements, subsystems, system and context.

- Learning coupling: Here, learning refers to the methods, objectives, tasks, levels, dimensions, processes, measures and learning outcomes associated with understanding a problem; accordingly, couplings exist within or between them. Typical examples include ensemble clustering [60] which considers couplings between clustering methods and within a clustering, and pattern combinations [5, 11, 12] which consider sub-patterns that are aggregated in terms of certain relations.

Limited work can be found in the literature on handling the above aspects of couplings. Addressing each of the above aspects or specific elements within an aspect has the potential to not only upgrade existing theories but also create new theoretical breakthroughs.

### 3.2. Coupling forms

Couplings in complex data and applications may have different forms, configurations and structures, which are often mixed with each other. Such couplings may need to be explored from structural, semantic, probabilistic, mathematical, dynamic, graphical/image, and/or networking perspectives.

Different forms of coupling relationships exist in complex applications such as behavioral and social systems [6]. Following our discussions about complex relations in complex behavioral systems in [58, 5], the following types of couplings may appear between the aspects discussed in Section 3.1. We highlight them because they are common and have been studied in several areas. In the example, we particularly illustrate such type of coupling that exists in behaviors within a social media system.

- *Serial coupling*: There is a serial order between adjacent entities; for instance, one behavior happens after another, one item is purchased after another, or one comment in a blog triggers another comment on the same topic. Two common types of serial couplings are prefix couplings and postfix couplings, in which one entity connects to another in either a prefix or postfix way. Serial coupling is particularly explored in sequence analysis [69] and behavior computing [13, 8, 4].

- *Causal coupling*: One entity causes the occurrence or dynamics of an effect or phenomenon as a consequence of another. For instance, a large buy order triggers significant market movement or a breaking news story, causing a significant increase in discussions on Twitter. Typical causal coupling learning tasks involve the learning of causation between factors (in particular entities and properties) and phenomena (such as anomaly).

- *Synchronous coupling*: Entities occur or contribute to an effect at the same time. For instance, several behaviors or social events occur at the same time, for example, two bloggers express similar comments on the same issue on different social media at the same time.

- *Exclusive coupling*: Entities take part or play roles in an exclusive way. For example, different events happen on a mutually exclusive basis; two opponent groups share different views on the same social event in a blog; or an attribute presents one particular value in a certain circumstance.

- *Dependent coupling*: The occurrence of one entity or entity status relies conditionally on another. For instance, the occurrence of a behavior relies on the pre-occurrence of a series of other behaviors. Typical learning tasks include hidden Markov model-based coupled behavior analysis [8] and copula-based financial variable dependence analysis [64, 7, 65, 66].

Different forms of couplings take place in different coupled objects. Couplings in numeric data are very different from those in categorical data. Of all the involved attributes, couplings include single entity-based couplings such as temporal coupling, pairwise coupling involving two entities, and compound coupling involving more than two entities, such as hierarchical coupling. From the knowledge representation aspect, syntactic coupling, semantic coupling, and inferential coupling can be explored. In data formation, coupling may be embedded within or between arrays, affinities such as associations, sequences, episodes, paths, artworks such as design patterns, mosaics, tessellations, and networks and graphs. In [58, 5], different temporal couplings and inferential couplings are discussed for coupled behaviors.

In addition, there are couplings on different levels, from value, attribute, object, method, and measure to pattern. Such couplings, which are more comprehensive and complex than correlation and association, refer to the relations that exist explicitly or implicitly between source and destination entities. A source or destination entity can be a value, attribute, object, method or pattern in a problem.

Often, we only focus on explicit couplings which are visible to us and easy to learn. Typical work in the hybridization of multiple methods and combining multiple sources of data into a big table for analysis fall in this category. Learning about implicit or invisible couplings is required in handling complex data and applications such as group behavior analysis [8]. Such couplings may take invisible forms and are thus hard to quantify. Data-driven learning of invisible relations is an important way to explore them.

In the related work, the above comprehensive couplings are often ignored or only partially addressed. Only a certain relation or correlation is considered. For example, in recommender systems, only user-user influence or item-item co-occurrence is considered. There are huge opportunities to invent new methods to explore the above different forms of couplings, especially implicit and invisible couplings in a learning system.

## 4. Learning coupling

Learning coupling refers to understanding, formalizing and quantifying the coupling aspects, entities, interactions, layers, forms and strength. This includes extracting, discovering and estimating the interactions and relationships between learning components, including method, objective, task, level, dimension, process, measure and outcome, especially when the learning involves multiples of one of the above components, for instance, multi-methods or multi-tasks. Recently, the concept of *combined mining* [11, 71, 72] was proposed to analyze the coupling relationships between data sources, features (or feature sets), methods and learning outputs, which form the tasks of multi-source combined mining, multi-feature combined mining, multi-method combined mining, and combined pattern mining respectively. Combined mining represents the effort of proposing general methodologies suitable for handling couplings between different aspects.

In general, one may only focus on analyzing couplings between entities and their properties (attributes). This constitutes the mainstream focus in areas including data mining and machine learning, for instance, the multi-source, multi-database, multi-view, multi-instance, multi-class, and multi-label analytical tasks. Another recent focus has been on engaging multiple methods for ensemble learning. Increasing attention has been paid to learning about the environment, constraints, interactions, and pattern relations [5, 12] in analyzing complex data and business. In this section, we discuss modeling couplings, measuring couplings and the curse of coupling.

### 4.1. Modeling couplings

Typically, we are concerned about learning couplings at a range of levels from values, attributes, objects, methods and measures to learning outcomes (patterns). Such couplings, which are more comprehensive and complex than correlation and association, refer to the relations that exist explicitly or implicitly between source and destination entities and between different aspects.

- Value coupling: Coupling between the values of a property is called *value coupling*, and is also called intra-attribute or intra-property coupling. In some cases, values are dependent; that is, dependency (or broad relations) between values exist. Numeric values share different coupling types and measurements from categorical values, therefore, different metrics need to be developed to measure value couplings. For instance, to calculate similarity between numeric values, Euclidean distance, Manhattan distance, cosine distance, and correlation distance are often used. For categorical (Boolean and string) data, metrics such as Hamming distance, Jaccard dissimilarity, matching similarity, and Edit distance are used. (See examples in Section 6.1.1, particularly [59] for calculating categorical value coupling and [61] for numerical value coupling.)

- Property coupling: Coupling between properties, also called *property coupling* (or attribute coupling), reflects the influence of one property on others. Couplings between different types of properties need to be considered in different ways, for instance, as categorical property couplings and numerical property couplings. The calculation of property coupling is built on value coupling. See the inter-attribute coupling similarity in [59, 61] for examples.

- Object coupling: Here, objects refer to a common type of entity, i.e. data objects in a data set, which share similarity and dissimilarity, forming *object coupling*. The measurement of object coupling is achieved through coupling calculation for object properties and values. A typical challenge is the heterogeneity between objects which may be considered separately in object coupling analysis. The similarity metrics reported in [59] are designed to analyze coupled categorical objects, while those in [61] is for coupled numerical objects. In another case, the term similarity in document analysis does not have the property level, and [17] shows methods for measuring term coupling.

- Method coupling: When multiple methods (which may also be called approaches, models or algorithms) are involved, different methods may only capture a partial picture of the underlying problem, and there may be an overlapping area of different methods. This provides good reasons for ensemble learning and hybrid intelligent systems. However, the coupling between methods is not usually considered. Accordingly, *method coupling analysis* considers the coverage, difference and complementarity between methods in learning a problem. For example, in [60] we report some preliminary results on incorporating the coupling between ensemble clusterings.

- Pattern coupling: Here, pattern refers to broad learning outcomes, including frequent patterns and outliers, classes and clusters identified. There may be a relationship between the patterns discovered by one method or multiple methods, forming *combined patterns* such as pair patterns with pairwise subpatterns, and cluster patterns consisting of multiple subpatterns, as discussed in [11, 5]. Couplings between patterns form the basis of the need for *pattern relation analysis* [5], which may be explored from semantic, structural and knowledge representation perspectives. In [32], relations between multiple rules are explored to identify those rules which optimally contribute to the most cost-effective detection of online banking risk.

Overall, it can be seen that modeling coupling is a very promising area for further exploration. Different theories and tools need to be built to model the above different types of couplings. Although it is hard to make them generic in capturing the diverse couplings discussed in Sections 3.1 and 3.2, here we discuss some general methodologies learned through our preliminary efforts in relevant areas (see corresponding references), which may be useful for modeling different types of couplings.

- If the data related to a problem can be converted into *an information table* [6] or multiple information tables connected with one another, then the table can be a useful tool to carry forward coupling learning. See Figure 2 for the information table showing various couplings in recommender systems.

- A concept of *intra-coupling may be defined to capture interactions within a target element*, such as an object, a property and a method in an information table, and business objects, such as for a user or an item in a recommender system. This captures the relations within a column or a row in an information table. For example, in Table A in Figure 2, arrows connecting different ratings given by different users on the same item can be treated as intra-item coupling.

- *Inter-coupling may be defined to catch relations between homogeneous elements*, such as between samples, users and items, and *between heterogeneous elements*, such as between data sources, variables and methods, and between users and items in a social media system. The inter-couplings attract the interactions across columns or rows in an information table. For example, in Table A in Figure 2, the arrows connecting different ratings across items by a user can be treated as inter-item coupling.

- *Cross-domain coupling* may be specified to record coupling (including interactions and influence) from one table to another. Here, each table refers to a particular domain. For instance, coupling across user Table B and item Table C in Figure 2 shows the relationship between users and items in recommendation.

- Depending on the nature of the problem, *hierarchical coupling* may be considered to seize the connections between system levels, for instance, from object to object group, and from subcategory to category, as shown in Table C in Figure 2.

- *Aggregated coupling* takes into account and combines all couplings, namely intra-couplings with inter-couplings, cross-domain couplings involving different domains, or hierarchical couplings rolling from a low level of couplings to higher levels of couplings. The example of coupled ensemble clustering in [60] shows a way of aggregating value coupling, property coupling, object coupling, cluster coupling and clustering coupling for coupled ensemble clustering.

## 4.2. Measuring couplings

In modeling couplings, a very fundamental issue is how to measure coupling strength. This involves the need to answer two challenging questions for different purposes:

(1) *How do I know whether couplings exist in a data set, to what extent, and in what forms, that is, how can coupled datasets be tested?* and

(2) *How can I measure the extent of coupling within or between specific aspects, as discussed in Section 3.1, and in particular forms, as discussed in Section 3.2?*

Answering both questions is actually very challenging, as the concept is very new to the respective research communities, including machine learning, data mining, information retrieval and even statistics. This needs to be explored from the data characteristic perspective. Our preliminary exploration of non-IIDness learning [6, 8, 59, 60, 61, 17] gives us the following pointers that may be useful in addressing the above research problems:

- In order to understand a data set, we need to understand its data structure, distributions, and main characteristics. If the data can be converted into an information table or multiple information tables, it is then reasonable to discuss the different types of couplings in Section 4.1. This exercise may help us to understand what couplings are in the data.

- Classic factor analysis and descriptive methods such as PCA may be used to analyze the features/variables and feature relation. This may help us to capture explicit and significant coupling before we seek to identify the much more challenging implicit coupling (which often contributes more to big data applications, as the blind person recognising the elephant metaphor or Simpson's Paradox [62] show).

- If the data follows a statistical distribution which has been well studied, one direction to pursue is to expand the corresponding similarity measures by incorporating visible and invisible couplings, as discussed in Sections 3 and 4.1.

- If the data is too complex to understand, it is then hard to test what the couplings are and how important they are. A comprehensive exploration of interactions in an information table related to intra-couplings, inter-couplings, cross-domain couplings, and hierarchical couplings would be extremely useful for extracting couplings by a data-driven approach.

- It may be necessary to quantify the level of couplings related to intra-couplings, inter-couplings, cross-domain couplings, and hierarchical couplings, by inventing new coefficient or similarity metrics, for example.

- If the data is complex and very domain-specific, the methodologies discussed in domain-driven data mining [14, 9] concerning data intelligence, domain intelligence, human intelligence, organizational intelligence, social intelligence and networking intelligence [9] may provide a general guide for extracting different forms of interactions related to data, domain, human, and organizational and social factors. This requires domain knowledge and the assistance of domain experts to specify and extract interactions known to them, and to provide hints about unknown interactions.

- Sampling a data set and then testing whether the data fits certain distributions or a mixture of several distributions is a way to explore the data structure and characteristics.

Once the data-driven approach works well, it is important to develop benchmarking data sets embedded with certain levels and types of couplings. This will lead to the development of baseline approaches.

Equally challenging is the exploration of the design for Question (2), which requires the definition of new coupling metrics to measure specific couplings.

- Correlation coefficient and distance-based similarity metrics are widely used for numeric couplings. These tools can be used to measure coupling between values, properties, objects, methods and patterns as shown in [17, 59, 61]. In other cases, they may be expanded to capture invisible coupling and coupling that cannot be represented in the preferred form.

- For categorical data, matching, instance-based overlapping, co-occurrence frequency etc., are typical paradigms for considering coupling measurement. The different focus here may be on how to use these tools to fully disclose the hidden coupling within and between elements.

- Real-life data, especially in a big data environment, often involves multi-sourced and multi-structured data sets. Exploring such data essentially requires multiple methods to be learned on multiple sources and multiple feature sets. In addition to the local issue of modeling coupling in each data source, how to match and align findings from different sources, or of different structure, with a global picture is a key problem. This may need to be considered beyond normalizing and combining individual metrics, rather than considering the coupling between the outcomes (involving pattern coupling).

- For an individual learning task, coupling can be measured in terms of the respective learning methods. When ensemble methods are involved, the overall similarity measure needs to consider both local coupling-based similarity between heterogeneous values, properties, objects and methods in a learning task, as well as coupling between several methods used for the same learning task.

## 4.3. Curse of coupling

While it is certainly important and valuable to incorporate couplings within and/or between entities, attributes, context, interaction and learning modules, the consideration of couplings in learning is a very costly and challenging process. As far as we can see at the moment (the complexity of the problem has not been thoroughly explored due to this issue being at such an early stage), there are several aspects that need to be considered in coupling learning.

First, it is often not clear to us what kinds of couplings are in the data, and we do not know which particular couplings we are unaware of. This is particularly the case, and very challenging, in such complex data and applications as big data-related business, which involves multiple business lines associated with different management systems and governance structures in different regions. In Section 5, Figure 2 shows different couplings in a recommender system, which could inspire the exploration of both explicit (Tables A, B and C) and implicit (Table D) couplings, and low-level (Tables B,C,D) and high-level (Table A) couplings.

Second, more couplings in learning may incur little additional gain but at a substantially higher cost. The design of similarity and objective function for coupling learning, such as the coupled clustering in [59] and the coupled behavior analysis in [8], needs to consider a certain level of tradeoff between computational cost and learning performance gain. For this, the aspects shown in Figure 1 which are relevant to coupling need to be carefully selected, filtered, constructed or mined.

In addition, in complex problems such as extremely rare fraud detection [32], the couplings between some elements may be much weaker than others; however, they may happen to be interesting. For instance, in online banking risk management, the ratio between fraudulent and genuine transactions may be 1:1,000,000. Those transactions and banking behaviors with strong couplings are not likely to be problematic. To handle such challenging problems, the learning models may focus on capturing abnormally weak couplings rather than strong ones, so the information table based coupling learning framework may not work well.

Last but not least, the coupling learning process will undoubtedly raise many unprecedented issues and questions about couplings and coupling learning: for instance,

- Where are the specific couplings?

- What kinds of couplings are there?

- How can different couplings be represented?

- How can various couplings be measured?

- How can couplings be incorporated into learning models (objective functions)?

- Would an objective function supporting coupling learning be substantially different from existing models?

- If different couplings need to be considered, how can they be aggregated at the global level?

- How can the computational costs be reduced to make coupling learning suitable for large scale data? and

- How can the performance of coupling learning be evaluated?

The exploration of these questions will certainly create new theories and tools for learning couplings in complex data and applications. This observation applies to different areas, including statistics, data mining, machine learning, information retrieval and general information processing.

## 5. An example: Couplings in recommendation

In recommender systems such as online shopping websites, online broadcasting systems, IPTV, and social media, there are different types of intrinsic interactions: user-user couplings, item-item couplings, and user-item couplings. A user's behavior may influence his/her friends, which further affects the behaviors of others. Item attributes such as item price and quantity are often associated with each other. The price of one item may affect the price of another. An item may influence the sale market of another. To ensure accurate recommendations, comprehensive couplings between item attributes, users, items, and between users and items need to be considered. In this section, we expand the concept of recommendation to particularly concern coupling relationships embedded in these aspects, which have not been thoroughly considered by the research community.

### 5.1. The recommendation problem

#### 5.1.1. A new perspective

Here, we use the toy example illustrated in Figure 2 to propose a general framework for discussing and learning couplings in recommender systems, a new perspective for recommendation learning. The recommender system problem in Figure 2 can be represented by learning from multi-sources of couplings. As shown in Figure 2, assume we have three data sources:

- Table A which records the preference (rating) $A_{b,c}$ of a user $b$ on item $c$;

- Table B which captures user demographic information, in which a user $b$ is represented by user demographic properties $p_1$ to $p_n$ such as name, sex, age and city;

- Table C which stores item properties $q_1$ to $q_m$ such as price, category and subcategory for each item $c$.

Further, we assume there are couplings between users and between items, as well as between users and items, thus the existing decoupling-based approaches such as collaborative filtering models [3] do not work well. The learning of comprehensive couplings in a recommender system on such a three-source coupled data can be described as follows.

- Preference couplings in Table A: reflecting the explicit and subjective interactions and preferences of a user for an item, which is the area mainly explored by the current recommender system community.

- User couplings in Table B: reflecting the interactions and influence of one user on another through both the internal factors of a user and external interactions such as friendship between users.

- Item couplings in Table C: indicating the relationships and connections between items through item properties and item-item connections.

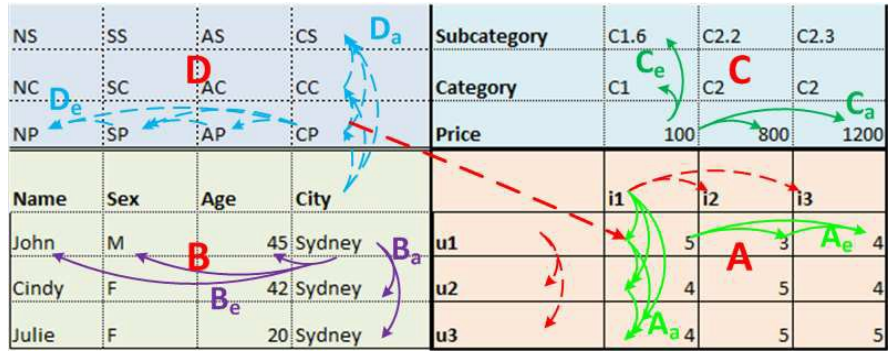

Figure 2. Couplings in recommender systems.

- User-item couplings in Table D: representing the implicit and objective interactions between user properties and item properties.

- Aggregated couplings: involving all couplings in Tables A, B, C and D, and combining both the explicit user-item couplings represented in preference couplings in Table A with the implicit user-time couplings in Table D, as well as the impact of user couplings and item couplings on user-item couplings for prediction.

### 5.1.2. Related recommendation research

Algorithms and methods for recommendation [28, 49] can be broadly categorized into four approaches: two basic approaches, namely collaborative filtering and content-based filtering, hybrid approaches and others.

Collaborative filtering (CF) predicts the rating a user may give to an item based on their own or other users' rating behavior. When other users' behavior is involved, a CF algorithm assumes the automatic collaboration of multiple users, and relies on group behavior and preferences similar to those of that particular user.

Content-based filtering (CBF) involves a user's comments on items, materials read or additional information such as the user's blogs. This content can be manually defined as needed.

Hybrid approaches include the integration of CF and CBF in various ways such as integrating models or modeling results to generate more accurate recommendation. The hybridization may take place in parallel (CF and CBF separately analyze the respective data and then the results are merged), or serially (e.g., CBF may be used to identify users with similar sentiments and then CF is employed to make recommendations on user behavior).

Metrics and tools used for measuring recommendation quality are consistent with the approaches used. The Pearson correlation is a popular algorithm for CF. Clustering algorithms such as K-Means or K-Modes are typical tools for grouping items that share similar rating behavior from users without any prior information about the grouping and similarity. Many other data mining and machine learning methods are applicable for recommendation; they include Fuzzy C-means, Adaptive Resonance Theory, probabilistic clustering (Expectation-Maximization), Bayesian Belief Nets, Markov chains, and Rocchio classification.

More recently, recommendation researchers have become interested in group recommendation [43] which focuses on group behavior analysis, and cross-domain recommendation [41, 42] which borrows insights gained from one domain to inform a decision on another.

So far, to the best of our knowledge, limited work has been done to incorporate couplings within and between ratings, users, items, and between users and items as shown in Figure 2. We undertook some preliminary work in [33] in this direction. Section 5.2 provides a general statement of these types of couplings for a new generation of recommendation research.

### 5.2. Couplings in recommender systems

We will elaborate on different kinds of couplings in a recommender system, as shown in Figure 2. We use the following specifications to define the diversified couplings:

. A capital letter $X$ refers to the entity set in a table in the figure, a lower case $x$ refers to a cell in the table; accordingly $X$ is instantiated to $A$ - ratings in Table A, $B$ - users in Table B, $C$ - items in Table C, and $D$ - user-item coupling in Table D. $X_{i,j}$ indicates a matrix showing the values related to two dimensions $i$ and $j$. For instance, $A_{i,j}$ refers to the rating matrix on users and items, and $a_{i_1,j_1}$ refers to the cell associated with user $i_1$ on item $j_1$.

. $X_a(.)$ (or simply $X_a$) refers to the intra-coupling within entities in a table. For instance, $B_a(\ )$ refers to the intra-coupling within user property values.

. $X_e(.)$ (or simply $X_e$) refers to the inter-coupling between entities in a table. For instance, $C_e(\ )$ refers to the inter-coupling between item properties.

. $X(.)$ represents a function for aggregating coupling in a table. For example, $C(\ )$ indicates all coupling existing in Table C.

· $X(X_a(\cdot), X_e(\cdot))$ describes the integration of intra-coupling $X_a(\cdot)$ with inter-coupling $X_e(\cdot)$ in terms of aggregation function $X(\cdot)$.

. $\mathsf{RS}^X$ presents the overall coupling similarity of entities $X$ in the respective table. $RS\ (.)$ refers to a function to integrate different aspects of coupling.

### 5.2.1. Preference couplings

Preference couplings, also called explicit or subjective user-item couplings, record the ratings given by a user to an item, which reflect the subjective preference of users for items. From a purely subjective perspective, there are user-user rating and item-item rating couplings in Table A. The rating couplings $A(\cdot)$ thus indicate the relationship between the ratings given by a user to an item, which can be categorized as intra-rating couplings between users $A_a(\cdot)$ and inter-rating couplings between items $A_e(\cdot)$, and the aggregated rating couplings $A(A_a(\cdot), A_e(\cdot))$.

**Definition 5.1.** *(Explicit user-item coupling) The explicit user-item coupling $RS^A$ refers to the coupling between preference rating $r_{i\,j}$, as shown in the rating Table A, each rating reflecting the subjective sentiment of a user $u_i$ on an item $i_j$. By incorporating the coupling between items and users in Table A, the overall explicit coupled user-item similarity $\mathsf{RS}^A$ can be measured by:*

$$\mathsf{RS}^A = A(A_a(\cdot), A_e(\cdot)) = A(A_{j_1,j_2}, A_{i_1,i_2}) \tag{1}$$

*where $A_{j_1,j_2}$ refers to rating coupling across $j_1$ and $j_2$ items (namely intra-coupling in Table A), and $A_{i_1,i_2}$ refers to rating coupling across $i_1$ and $i_2$ users (the inter-coupling in Table A).*

*Alternatively, if the non-IIDness in Table A can be or is actually overlooked, then*

$$\mathsf{RS}^A = A(A_{i,j}) \tag{2}$$

*where $A_{i,j}$ represents the preference rating matrix, in which $A_{i_1,j_1}$ is the rating of user $u_{i_1}$ on item $i_{j_1}$, and $A(.)$ represents the aggregation function. The ratings disclose an empirical, direct and explicit indication of the couplings between users and items.*

### 5.2.2. User couplings

Users are coupled. The user Table B consists of intra-user couplings $B_a(\ .)$, inter-user couplings $B_e(\ )$ and the aggregative couplings $B(B_a(.), B_e(\ ))$. We say users are coupled, hence they are called *coupled users*. The overall coupling $\mathsf{RS}^B$ in Table B can then be represented as:

$$\mathsf{RS}^B = B(B_a(\cdot), B_e(\cdot)) \tag{3}$$

### 5.2.3. Item couplings

Similarly, items are coupled. The item Table C consists of intra-item couplings $C_a(.)$, inter-item couplings $C_e(\ )$, and the aggregative couplings $C(C_a(.), C_e(\ ))$. We say items are coupled, hence they are called *coupled items.* Accordingly, the coupling in Table C is $\mathsf{RS}^C$:

$$\mathsf{RS}^C = C(C_a(\cdot), C_e(\cdot)) \tag{4}$$

### 5.2.4. User-item couplings

Since users and items are coupled respectively and an item is associated with one to many users, users and items are coupled too. This user-time coupling can be described in terms of two types:

· *explicit user-item couplings* (or *subjective user-item couplings*) in Table A as discussed in Section 5.2.1 and

· *implicit user-item couplings* (or *objective user-item couplings*) in Table D between users and items.

Although the ratings reflected in Table A are subjective, the overall rating behavior may be well supported by underlying reasons. Such underlying driving factors may be explored from objective and implicit perspectives, reflected by the underlying implicit interactions between users (coupled users) in Table B, items (coupled items) in Table C, and between users and items (which we call *implicit user-item coupling*) in Table D. In particular, any coupling function for a specific cell such as $s_{i_1 j_1}$ in Table D is the product of matrix $C_a$ for specific item property $q_{j_1}$ and matrix $B_a$ for specific user property $p_{i_1}$. We call this a *coupled user-item cell.* The coupling in each cell in Table D, namely $D_{i_1 j_1}$, may be a matrix to learn.

**Definition 5.2.** *(Coupled User-Item Cell) A coupled user-item cell $d_{ij}$ corresponds to all users and items which satisfy a specific user $i$'s property and an item $j$'s property.*

For example, in Table D in Figure 2, a cell CP captures all users and items with the coupling between the *City* property in user Table B and the *Price* in Table C for all users and items.

**Definition 5.3.** *(Implicit user-item cell coupling) Implicit coupling in a user-item coupled cell $D_{i_1 j_1}$ is $\mathsf{RS}^{D_{i_1 j_1}}$, which consists of two parts: the coupling of a user $i_1$'s specific property on all items with item property number $j_1$, namely $D_a(D_{i_1 j_{1*}})$ $(1 \le j_{1*} \le J)$; and the coupling of an item $j_1$'s specific property on all users with the user property number $i_1$, namely $D_e(D_{i_{*} j_1})$ $(1 \le i_{1*} \le I)$. As a result, the implicit coupling in a user-item coupled cell is $\mathsf{RS}^{D_{i_1 j_1}}$:*

$$\mathsf{RS}^{D_{i_1 j_1}} = RS^{D_{i_1 j_1}}(D_a(D_{i_1 j_{1*}}), D_e(D_{i_{1*} j_1})) \tag{5}$$

**Definition 5.4.** *(Implicit user-item coupling) The overall implicit user-item coupling hidden in Table D has the similarity $\mathsf{RS}^D$, which is the aggregation of all user-item cell coupling in Table D:*

$$\mathsf{RS}^D = RS^{\,D}(\mathsf{RS}^{D_{i_1 j_1}}) \tag{6}$$

*where $(i_1\ i_2) \lor (j_1\ j_2) \land (1 \le i_1, i_2 \le I) \land (1 \le j_1, j_2 \le J$. Alternatively, we may say the overall implicit user-item coupling consists of two parts: a user $i_1$'s property-based item coupling $D_a(\cdot)$ on all item properties, and an item $j_1$'s property-based user coupling $D_e(\cdot)$ on all user properties, which are further coupled in terms of $RS^{\,D}$.*

$$\mathsf{RS}^D = RS^{\,D}(D_a(\cdot), D_e(\cdot)) \tag{7}$$

where $D_a(.)$ represents the coupling between different item properties on a particular user property, and $D_e(\ )$ indicates the coupling between user properties in terms of a specific item property.

An example in Table D in Figure 2 is the cell CP. The full coupling in CP consists of $CP_a$ which captures all coupling between all item prices in a user's City, and $CP_e$ which calculates the coupling of all user cities on item property Price.

### 5.2.5. Aggregated coupling

We have discussed above the explicit user-item couplings embodied in Table A and the implicit user-item couplings in Table D. These only reflect partial aspects of the elements that contribute to the prediction of user ratings on items.

**Definition 5.5.** *(Aggregated User-Item Couplings) The aggregated user-item coupling is the combination through an aggregation function* $RS^{A+D}(\cdot)$ *of explicit user-item couplings* $\mathsf{RS}^A$ *and implicit user-item couplings* $\mathsf{RS}^D$:

$$\mathsf{RS}^{A+D} = RS^{A+D}(\mathsf{RS}^A, \mathsf{RS}^D) = \sum_{i_1,j_2=1}^{I} \sum_{j_1,j_2=1}^{J} RS^{A+D}(A(A_{i_1 j_1}, RS^D(RS_{i_1 j_1}^D))) \odot (A_{i_1 j_1}, RS_{i_1 j_1}^D) \tag{8}$$

where $RS^{A+D}(A(A_{i_1 j_1}, RS^D(RS_{i_1}^D)))$ means the subsequent coupling of $RS^{A+D}$ are $A_{i_1 j_1}$ coupled with $A(A_{i_1 j_1'}, RS^D(RS_{i_1}^D))$, and $RS_{i_1}^D$ coupled with $RS^D(RS_{i_1}^D)$, and so on, with non-determinism.

The objective of a recommender system is to predict a particular rating $\hat{r}$ by a user on an item, which results in a rating matrix $\hat{R}$. For this, the complete coupling in a recommender system is embodied through four sources: coupled users in Table B, coupled items in Table C, explicit user-item interactions in Table A, and implicit user-item interactions in Table D.

**Definition 5.6.** *(Complete Coupling) The complete coupling* $\mathsf{RS}$ *in a recommender system requests the aggregation of the actual preference* $\mathsf{RS}^A$ *of a user on an item in Table A, an accumulative indication* $\mathsf{RS}^B$ *of the influence between users determined by user properties and its impact on prediction, the relation between items* $\mathsf{RS}^C$ *shown by item properties, and the underlying hidden yet comprehensive interactions* $\mathsf{RS}^D$ *between the user properties and item attributes driving the rating.*

$$\mathsf{RS} = RS(\mathsf{RS}^A, \mathsf{RS}^B, \mathsf{RS}^C, \mathsf{RS}^D) \tag{9}$$

Note that $\mathsf{RS}^D$ is not a simple matrix, because it carries much information from Tables B and C and their interactions on many different properties, at different layers and on various forms. It is in fact very complicated to obtain $\mathsf{RS}^D$, so future studies must incorporate the implicit interactions between properties in Tables B and C.

The above discussions about diverse couplings in a recommender system indicate a new paradigm which shifts the focus of recommendation design from understanding explicit user-item preference to considering both explicit and implicit user-item interactions across multiple sources, with different properties, layers and forms. This will trigger research into the next generation of recommendation algorithms and systems.

## 6. Case study: Coupled recommender systems

The discussions about learning different types of couplings in recommender systems in Section 5 inspire us to incorporate couplings into recommendation algorithms. In this section, we discuss two preliminary studies in this direction. The first (for more details, see [70]) considers coupled item recommendation, which incorporates couplings into items and creates a new coupled collaborative filtering (CCF) algorithm: Coupled K-modes (CK-modes). The second (for more details, see [33]) considers the couplings between users and between items and proposes a new coupled matrix factorization (CMF) algorithm. Readers who are interested in these can find more detail in [33, 70].

### 6.1. Coupled item recommendation

As discussed in [6], the classic collaborative filtering algorithms ignore or only partially consider the couplings between item properties, user properties, and item-user interactions. Here, we present a coupled item-based CF by explicitly considering both intra-coupling and inter-coupling between item attributes, and aggregating them in terms of the coupled object similarity (COS) proposed in [59].

### 6.1.1. Measuring item attribute couplings

The coupled item similarity (*CIS*) between the categorical data type of items $x$ and $y$ is defined as follows:

$$CIS(x, y) = \sum_{j=1}^{n} \delta_j^A(x_j, y_j), \tag{10}$$

where $x_j$ and $y_j$ are the values of item attribute $j$ for items $x$ and $y$, respectively; $n$ refers to the number of item attributes; and $\delta_j^A$ is the Coupled Attribute Value Similarity (CAVS) (see [59] for details).

The *CAVS* is further described by the *Intra-coupled Attribute Value Similarity (IaAVS)* measuring the item attribute value similarity by considering the feature value occurrence frequencies within an item feature, and the *Inter-coupled Attribute Value Similarity (IeAVS)* measuring the item attribute value similarity by taking the item attribute dependency aggregation into account.

For item attribute $j$, *IaAVS* $\delta_j^{Ia}(x_j, y_j)$ is calculated as below [59],

$$\delta_j^{Ia}(x_j, y_j) = \frac{|g_j(x)| \cdot |g_j(y)|}{|g_j(x)| + |g_j(y)| + |g_j(x)| \cdot |g_j(y)|}. \tag{11}$$

where $g_j(x)$ returns all objects with values $x$ for feature $j$. *IeAVS* $\delta^{Ie}(x_j, y_k)$ is calculated as per

$$\delta_{jk}^{P}(x, y) = \min_{W \subseteq V_k} \{2 - P_{k|j}(W|x) - P_{k|j}(\overline{W}|y)\}. \tag{12}$$

where $\overline{W} = V_k \setminus W$ is the complementary set of a set $W$ under the complete set $V_k$. For the $k^{th}$ attribute value subset $W \subseteq V_k$, and the $j^{th}$ attribute value $x \in V_j$, $P_{k \in j}(W x)$ is the Information Conditional Probability (ICP) of $W$ with respect to $x$,

$$P_{k|j}(W|x) = \frac{|g_k^*(W) \cap g_j(x)|}{|g_j(x)|}. \tag{13}$$

where $g_k^*(W)$ maps the value set $W$ of feature $k$ to the dependent object set.

Accordingly, *CAVS* $\delta_j^A$ between item attribute values $x_j$ and $y_j$ of item attribute $j$ is defined as follows.

$$\delta_j^A(x_j, y_j) = \delta_j^{Ia}(x_j, y_j) \cdot \delta_j^{Ie}(x_j, y_j) \tag{14}$$

### 6.1.2. Coupled K-modes

Taking the K-modes clustering algorithm as an example, we create a coupled K-modes (CK-modes for short). Let $S$ be a cluster generated by the previous partition of the K-modes algorithm. In the cluster $S$, there are $J$ items described by categorical item features $a_{j_1}$, $a_{j_2}$, ... $a_{j_l}$. A mode of the cluster $S$ is an item vector $Q = [q_1, q_2, ..., q_J]$ to maximize the sum of the similarity $Sim(Q, S)$ between each element $i$ of $S$ and $Q$.

$$Sim(Q, S) = \sum_{i=1}^{J} CIS(S_i, Q) \tag{15}$$

Within the CK-modes model, the item-based collaborative filtering is adjusted to generate the prediction on item $y$ for an active user $u$ based on all other items $x$. The prediction $\hat{r}_{u,y}$ on item $y$ for an active user $u$ is computed by the following formula:

$$\hat{r}_{u,y} = \overline{r_{u,x}} + \frac{\sum_{x \in N} Sim(x, y) * (r_{u,y} - \overline{r_y})}{\sum_{x \in N} Sim(x, y)} \tag{16}$$

where $N$ is the number of joint items both rated by user $u$ and grouped by the CK-modes algorithm, and $r_{u,y}$ represents the rating on item $y$ given by user $u$. $Sim(x, y)$ is the coupled item similarity between items $x$ and $y$. $r_{u,x}$ is the average of active user $u$'s ratings.

### 6.1.3. Experiments

We evaluate CK-modes against several widely discussed algorithms in the recommender systems, including the user-based collaborative filtering algorithm [47], item-based collaborative filtering algorithm [52] and *CLUSTKNN* [1] on the MovieLens data. Figure 3 shows the *throughput* of all algorithms. Here, throughput represents the number of recommendations generated per second. The user-based recommendation algorithm scans the whole user-item matrix $R$, and its throughput does not change with the number of clusters. However, the *throughput* of the item-based recommendation algorithm varies with the number of neighbors selected for prediction. We plot the throughput of the item-based recommendation algorithm by setting the number of neighbors as 30, since this generates the best prediction quality.

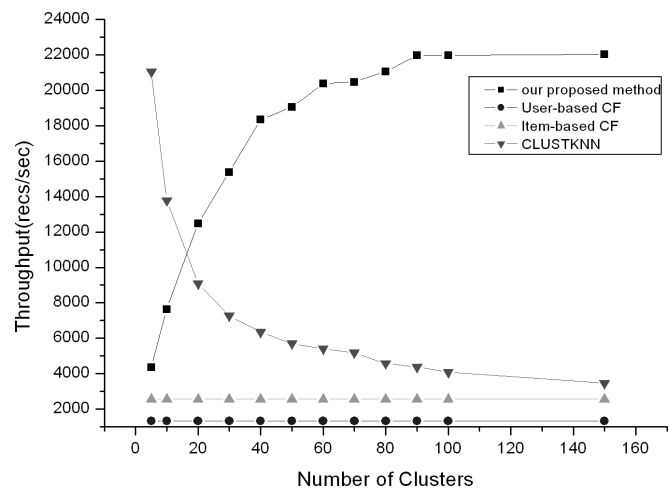

Figure 3. Throughput of the selected recommendation algorithms

This case study shows the potential of (and need for) incorporating coupling between item properties into item similarity analysis. Due to the additional information on the similarity between the values of an item attribute and between item attributes, and their aggregation brought to the similarity analysis, the implicit coupling relationships between items are extracted and explicated. Such coupling complements user rating behavior because it discloses the reasons for a user's preference for items; in other words, item attributes and attribute values drive user rating behavior.

Clearly, more work can be done on user-based collaborative filtering (UCF) to incorporate the user's driving forces (about user property and property values) into the user's rating behavior on items. More work can be done on coupled user-item collaborative filtering (CCF) to incorporate both user-coupling and item-coupling to extract and explicate the hidden coupling relationships and factors that drive a specific user's preference for particular items.

### 6.2. Coupled matrix factorization

The basic matrix factorization approach for recommendation builds on an assumption that users are independent and identically distributed (IID). An MF algorithm [31] is a latent factor model which is generally effective for estimating the overall structure that relates simultaneously to most or all items.

This approach ignores the social activities between users, which is inconsistent with the reality that we normally ask friends for references or follow trusted recommendations. This fact is increasingly respected in social network and social media study. Social recommender systems and a variety of models have been proposed, which engage social network analysis with recommender systems, forming approaches such as Social Recommendation (SoRec) [38], Social Trust Ensemble (STE) [39], and recommender systems with social regularization. Social Matrix Factorization approaches consider the social activities of users as a group, in which the social relations between users are mixed together and treated equally. As a result, it is impossible to differentiate social recommendations from different friends in terms of their personalized preferences and influence. Apart from this, item intra-couplings are also ignored.

### 6.2.1. Basic matrix factorization model

The basic matrix factorization works as follows. The matrix of predicted ratings $\hat{R} \in \mathbf{R}^{u_0 \times i_0}$ is modeled as:

$$\hat{R} = R_m + PQ^T \tag{17}$$

with matrices $P \in \mathbf{R}^{u_0 \times d}$ and $Q \in \mathbf{R}^{i_0 \times d}$, where $u_0$ denotes the number of users, $i_0$ is the number of items, $d$ is the rank (or dimension of the latent space) with $d \ll i_0$, $u_0$, and $R_m \in \mathbf{R}$ is a global offset value. $PQ^T$ reflects the interaction between users and items. Through a gradient descent approach, the best matrices $P$ and $Q$ can be computed.

Equation (17) shows that the prediction by matrix $\hat{R}$ is transferred to compute the mapping of items and users to factor matrices $P$ and $Q$. Once this mapping is built, the model predicts the rating a user will give to any item by Equation (17). The computation of the mapping can be optimized by minimizing the regularized squared error on the set of known ratings, as shown in Equation (18) [68].

$$L = \frac{1}{2} \sum_{(u,i) \in K} \left( R_{ui} - \hat{R}_{ui} \right)^2 + \frac{\lambda}{2} \left( \|Q_i\|^2 + \|P_u\|^2 \right) \tag{18}$$

### 6.2.2. Incorporating couplings into matrix factorization

The basic MF algorithm directly computes the production of vectors $P_u$ and $Q_i$. However, the coupling relations within users and items have not been considered. In reality, not only do users interact with and influence each other, items are also more or less related due to item properties or usage purposes. Such user associations and item relations are not sufficiently considered in the above basic MF.

Here, we consider the low-level interactions including intra-couplings within users and items and inter-couplings between users and items through the preference rating table (Table A in Figure 2) only. Based on a user's interest in items, intra-couplings indicate the involvement of friendship between users and similarity between items, so inter-couplings stand for the association between users and items. Note that here, we do not consider interactions between users and between items from the perspective of their specific properties.

A new coupled MF model is generated below by incorporating the above item/user intra-couplings and user-item inter-couplings:

$$\hat{R}_{u,i} = R_m + \delta_{u,i}^{Ie} + \delta_{u,i}^{Ia} \tag{19}$$

where $\delta_{u,i}^{Ia}$ represents the intra-couplings within users and items, and $\delta_{u,i}^{Ie}$ represents the inter-coupling between users and items. Here, we assume there is a linear relationship between intra-couplings and inter-couplings in a recommender system.

Inter-couplings: Users choose or comment on the items they like or dislike. This is reflected by the preference ratings showing the intuitive association between a user and item(s), for instance, an author cites a paper. Here, we only consider the explicit user-item coupling (see Section 5.2.4), which is reflected by $\hat{\delta}_{i,i}^{Ie} = P_u Q_i^T$.

Intra-couplings: Intra-couplings here include user-intra-couplings and item-intra-couplings, for instance, an author collaborates with another, and a paper is cited by other papers. Assuming the rating of a user on items relies on the interactions between a user and other users, and similarly the influence of an item on all others exists in the rating, Equation (20) models such couplings:

$$\delta_{u,i}^{Ia} = \sum_{v \in N(u)} S_{u,v} P_v Q_i^T + \sum_{j \in N(i)} W_{i,j} P_u Q_j^T \tag{20}$$

where $N(u)$ and $N(i)$ refer to the number of users and items respectively; the first part is intra-user coupling, where $S_{u,v}$ reflects the friendship or social relation between users $u$ and $v$, and $W_{i,j}$ indicates the relation between items $i$ and $j$, which indicates, if item $i$ links item $j$, the preference of a user $u$ on an item $i$ would be passed to item $j$.

Equation (20) indicates that a user's profile $P_u$ is somehow similar to his/her friend's profile $P_v$, and if a user $u$ is interested in an item $i$, he/she may also be interested in item $j$ which is similar to $i$.

Accordingly, in [33], we propose a Coupled Group MF algorithm (CGMF) which also considers user grouping in social media to cater for a specific group profile on top of incorporating the above discussed intra and inter-couplings.

### 6.2.3. Experiments

We use MovieLens, LastFm and DBLP. The MovieLens 10M data set consists of 10 million ratings and 100,000 tag applications applied to 10,000 movies by 72,000 users. LastFm contains the social networking, tagging, and music artist listening information of 1892 users, 17632 artists, 12717 bi-directional user friend relations, 92834 user-listened artist relations, and 11946 tag assignments. The DBLP citation database contains 1,572,277 papers and 2,084,019 citations. Each paper is associated with title, abstract, authors, year, venue, citation number and references. The data set consists of various coupling relations such as co-authoring relations, citation relations and "write", or "write-by" relations.

In our experiments, the MovieLens and LastFm data sets are separately used for movie and artist recommendation, and the DBLP data set is explored for paper recommendations to test our CGMF model. In the group formation step, a paper's title and abstract information in the DBLP data set are explored to extract topics by LDA. Likewise, tags are used to group the artists and users in the LastFm data set. Different from DBLP and LastFm, the MovieLens's genre attributes are simply used to group the movies.

Five-fold cross validation is performed in our experiments. In each fold, we have 80% of data as the training set and the remaining 20% as the test set. Here, we use Root Mean Square Error (RMSE) and Mean Absolute Error (MAE) as the evaluation metrics.

$$RMSE = \sqrt{\frac{\sum_{(u,i)|R_{test}} (r_{u,i} - \hat{r}_{u,i})^2}{|R_{test}|}} \tag{21}$$

$$MAE = \frac{\sum_{(u,i)|R_{test}} |r_{u,i} - \hat{r}_{u,i}|}{|R_{test}|} \tag{22}$$

where $R_{test}$ is the set of all pairs $(u, i)$ in the test set, $r_{u,i}$ refers to a rating made by user $u$ on item $i$, and $\hat{r_u}_{,i}$ refers to the predictive rating.

To evaluate the performance of our proposed CGMF, we consider three baseline approaches:

- CF: This is an item-based collaborative filtering method called Slope One.

- BaseMF [51]: This method is a probabilistic matrix factorization approach which does not take social networking into account.

- SocialMF: This is the model described in Equation (18).

The effectiveness of each method on different data sets is shown in Figures 4, 5 and 6. Our proposed CGMF method outperforms the baselines in terms of RMSE and MAE. Experimental results on the DBLP data set (Figure 4) show that CGMF and SocialMF perform much better than CF and Basic MF. Compared to SocialMF, CGMF achieves up to 4.3% improvement on RMSE, 14.7% on MAE. On LastFm and MovieLens, Figures 5 and 6 show that CGMF also performs better than the others. This indicates that there is a possibility of improving the recommendation quality by incorporating item and user couplings into recommender systems.

This case study shows the value of and further opportunities for extracting and explicating user coupling and item coupling into matrix factorization to form a coupled matrix factorization (CMF). A CMF is more powerful because it captures the hidden coupling between users and between items represented by their properties and the coupling relationship between them. Such low-level coupling embodied between user (and item) property values and between properties serves as the driving force for a user's preference for particular items. The study also shows that, while classic MF captures the latent variables related to users and items, the above discussed coupling relationships are not necessarily captured in the classic MF model. Thus, the study of incorporating comprehensive coupling relationship into MF to generate a CMF is very necessary, especially for data with strong visible or invisible interactions.

## 7. Discussions

Coupling learning is a very promising direction in learning complex relationships between objects, properties, processes, facts, events, and states of affairs which are beyond correlation, association and dependency. Complex couplings are a major characteristic of big data, and together with heterogeneity form the phenomenon of non-IIDness, namely non-independent and non-identically distributed characteristics. Non-IIDness greatly challenges the exist- ing theories and systems in statistics, data mining, machine learning, text mining, information retrieval and pattern recognition, since the main theoretical foundation in these modern fields has been built on the assumption that the studied objects (as well as variables, processes, facts, events, and states of affairs) are IID. This is, in fact, a very strong assumption about or abstraction of complex real-world problems. The world has developed to the stage where non-IIDness needs to be tackled directly, thus new theoretical breakthroughs need to be pursued.

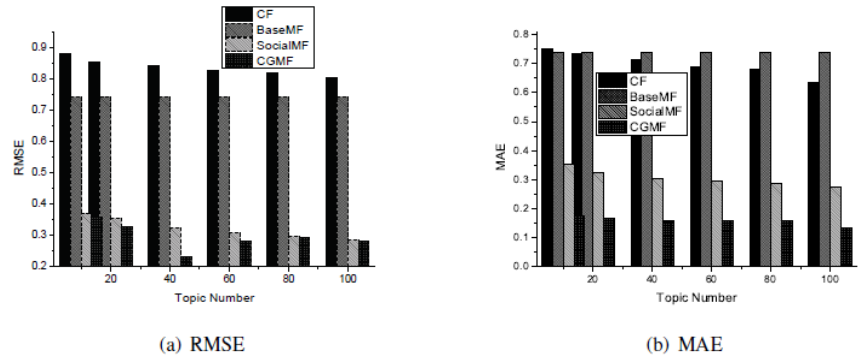

(a) RMSE  (b) MAE

Figure 4. RMSE and MAE Comparison on DBLP with a Different Number of Topics

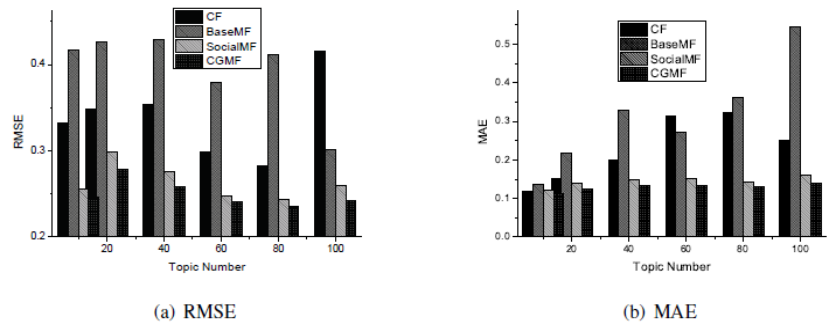

(a) RMSE  (b) MAE

Figure 5. RMSE and MAE Comparison on LastFm with a Different Number of Topics

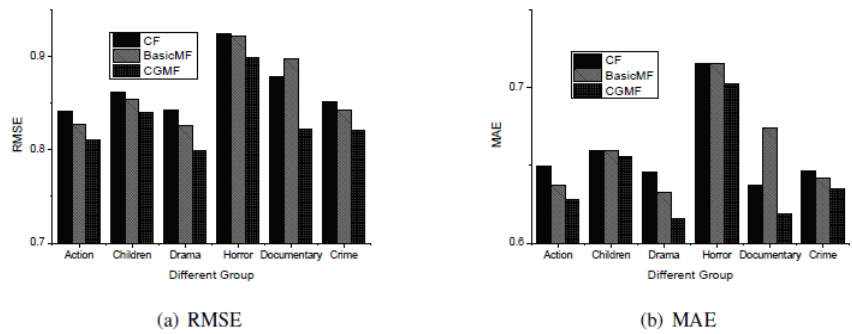

(a) RMSE  (b) MAE

Figure 6. RMSE and MAE Comparison on MovieLens with Different Groups

In our team, non-IIDness learning has become a major research direction in recent years, in collaboration with major government and industrial partners' data and business problems, which are certainly non-IID. We have demonstrated the effectiveness and value of exploring non-IIDness to advance data mining, machine learning and behavior informatics. In addition to the two case studies presented in Section 6, here we discuss several such exercises.

(1) Coupled clustering. In [59], a coupled K-modes (CK-modes) algorithm was designed to cluster categorical data, which caters for couplings within an attribute (intra-attribute coupling), between attributes (inter-attribute coupling), and between objects (coupled object similarity). This CK-modes algorithm has been successfully used in item recommendation, as discussed in Section 6.1 [70]. In [61], couplings within and between continuous variables are incorporated into learning, which has been shown to be effective for data clustering and classification by two proposed algorithms: Coupled Linkage algorithm (LC-CR) and Coupled Spectral Clustering (SC-CR), comparing to those without considering couplings.

(2) Coupled text analysis. Term-term couplings are important factors to be considered in text analysis which presents as direct or indirect relationships; for instance, semantic connections between terms across different pages in a paper or across different papers. In [17], the idea of couplings was introduced to analyze intra-term and inter-term coupling relationships to capture indirect semantic linkage, such as between "data mining" and "machine learning". An algorithm called Coupled Term-term Relation Model (CRM) was proposed and tested and found to be more effective than typical baselines in text analysis. The concepts and ideas introduced in this work can be extended to coupled information retrieval analysis and text mining by incorporating explicit and implicit (semantic or indirect) couplings between and within the text and information concerned.

(3) Coupled ensemble learning. Couplings between different learning methods are often ignored, or replaced by simple ranking or weighting-based aggregation. In [60], we demonstrate the need for and effectiveness of incorporating couplings within and between clustering methods, called intra-clustering and inter-clustering couplings, and within and between clusters (called intra-cluster and inter-cluster couplings), and coupled objects into ensemble clustering. The algorithm Coupled Clustering Ensemble (CCE) was shown to be more effective than the state-of-the-art algorithms.

(4) Coupled behavior analysis. Behaviors of individuals and groups are more or less related for temporal, socio-cultural and other reasons. Such relationships are embodied through intra-behavior couplings for one actor and inter-behavior couplings between multiple actors. Coupled behaviors exist widely in business and social events and problems, including transport systems, social media and networks, and financial markets. In [8], we formalize the problem of coupled behavior analysis, and propose a Coupled Hidden Markov Model (CHMM) and hybrid coupling and hierarchical coupling-based Relational Dependency Network to model conditional probability distributions to capture such couplings.

(5) Pattern relation analysis. Usually, data miners extract so-called patterns that satisfy certain criteria. There are different kinds of relationships between discovered patterns. This raises the issue of analyzing the relationships between patterns, which has been little explored in the data mining and machine learning community. The analysis of pattern relations is important for many reasons. For instance, a pattern represents the behavior modes of certain objects, thus the pattern relation involves the behavior relations between different groups of objects. In [5], we explore the issue of pattern relation analysis and different types of pattern relations. In [12], several new types of patterns were identified, including pair patterns, cluster patterns, and approaches to quantify combined patterns such as combined association rules [71, 72]. In [32], the interactions between rules are analyzed to avoid the duplication caused by multiple rules with overlapping functions while maintaining the detection quality of rule-based risk management.

(6) Testing coupled data. It is often asked what kind of data is coupled and can be used for coupling learning. In fact, real-world data sets are all embedded with more or less couplings. In our practice, we have tested couplings in real-world business, including capital markets [10, 8, 54], online banking business [32], recommender systems including social media [33], and web text [17]. We have shown that even a very commonly used data set, or the highly manipulated UCI data, can be used for coupling learning evaluation, as we showed in [59, 60, 61] for coupled clustering and coupled ensemble clustering, although the couplings may not be very strong, and the performance difference by incorporating couplings may not be very obvious.

Coupling learning directly contributes to non-IIDness learning [6], which is a fundamental issue facing existing learning, analytic, measurement and computational systems. The research on coupling learning essentially has great potential and offers opportunities for advancing the existing methodologies, algorithms, tools and systems in statistics, machine learning, data mining, computational intelligence, information retrieval, and broad information processing to the next generation.

## 8. Conclusions

In the real world, diverse coupling relationships are embedded in every business and are associated with objects, properties, processes, events, and states of affairs. Such couplings may present characteristics, which are far beyond the association, correlation and dependency relationships that usually concern statistics, data mining and machine learning communities. Triggered by behavioral, economic, social, cultural, or other driving forces, they may be explicit vs. implicit, syntactic vs. semantic, local vs. global, specific vs. comprehensive, or subjective vs. objective. Learning such couplings is far more complex than widely-explored association, correlation and dependency. The fact may be that we do not have a full picture of what kinds of couplings are in a system, how they present and evolve, and how to represent, model and measure them in a learning system.

In this paper, we have conducted a high level discussion about coupling learning. Our purpose is to create a high level picture of what couplings are, why they are important, how they may present, and the issues in modeling and measuring ubiquitous couplings from a generic and comprehensive perspective. We have demonstrated several examples and case studies showing what couplings may exist and how to specify them in a recommender system, as well as illustrating how such couplings can be modeled to improve learning objectives. The discussions and exploration have hopefully highlighted promising opportunities and prospects for incorporating couplings into advancing existing learning systems, and developing the next-generation learning, analytic, computational, and measurement theories and tools.